\documentclass[10pt,twocolumn,letterpaper]{article}

\usepackage{cvpr}
\usepackage{booktabs}
\usepackage{times}
\usepackage{epsfig}
\usepackage{graphicx}
\usepackage{amsmath}
\usepackage{amssymb}
\usepackage{multirow}
\usepackage{adjustbox}

\newcommand{\specialcell}[2][l]{%
  \begin{tabular}[#1]{@{}l@{}}#2\end{tabular}}

\usepackage[pagebackref=true,breaklinks=true,letterpaper=true,colorlinks,bookmarks=false]{hyperref}

\cvprfinalcopy 

\def\cvprPaperID{17} 

\ifcvprfinal\fi
\begin{document}

\title{Group Activity Detection from Trajectory and Video Data in Soccer}

\author{Ryan Sanford\thanks{Equal contribution.}~~~~Siavash Gorji\footnotemark[1]~~~~Luiz G. Hafemann\footnotemark[1]~~~~Bahareh Pourbabaee~~~~Mehrsan Javan\\[1mm]
\normalsize Sportlogiq, Montreal, Canada\\
{\tt\normalsize \{ryan.sanford, siavash, luiz, bahar, mehrsan\}@sportlogiq.com}
}

\maketitle

\begin{abstract}

Group activity detection in soccer can be done by using either video data or player and ball trajectory data.
In current soccer activity datasets, activities are labelled as atomic events without a duration. Given that the state-of-the-art activity detection methods are not well-defined for atomic actions, these methods cannot be used. 
In this work, we evaluated the effectiveness of activity recognition models for detecting such events, by using an intuitive non-maximum suppression process and evaluation metrics.
We also considered the problem of explicitly modeling interactions between players and ball. For this, we propose self-attention models to learn and extract relevant information from a group of soccer players for activity detection from both trajectory and video data.
We conducted an extensive study on the use of visual features and trajectory data for group activity detection in sports using a large scale soccer dataset provided by Sportlogiq.
Our results show that most events can be detected using either vision or trajectory-based  approaches with a temporal resolution of less than $0.5$ seconds, and that each approach has unique challenges. 
\end{abstract}

\section{Introduction}
Group activity recognition aims to understand the action of each individual and how they interact with each other in a group setting \cite{wu_learning_2019,ibrahim_hierarchical_2016,QiECCV2018, AzarCVPR2019,tran_learning_2015,ji_3d_2013}. The best examples of group activities are sports games, where a group of individuals are interacting with each other along with an object (e.g. 22 people in a soccer game or 12 people in a volleyball game). 
Detecting group activity in sports has several practical applications, such as assessing team strategy and players performance as well as providing relevant content to media~\cite{Schulte2017DMKD,Schulte2017SSAC,Keane2019SSAC}.


While visual data, such as videos and images, have been extensively used in the past for group activity recognition in sports \cite{roy_tora_classification_2017, fani_hockey_2017, cai2019temporal, piergiovanni2018fine}, several sports have player and ball/puck location data available. These are obtained from non-vision sensors, such as RFID transmitters, GPS trackers on players or extracted from video data. In this paper, we analyzed the use of both video and trajectory data for group activity detection in sports. 

\begin{figure}
    \centering
    \includegraphics[width=0.9\columnwidth]{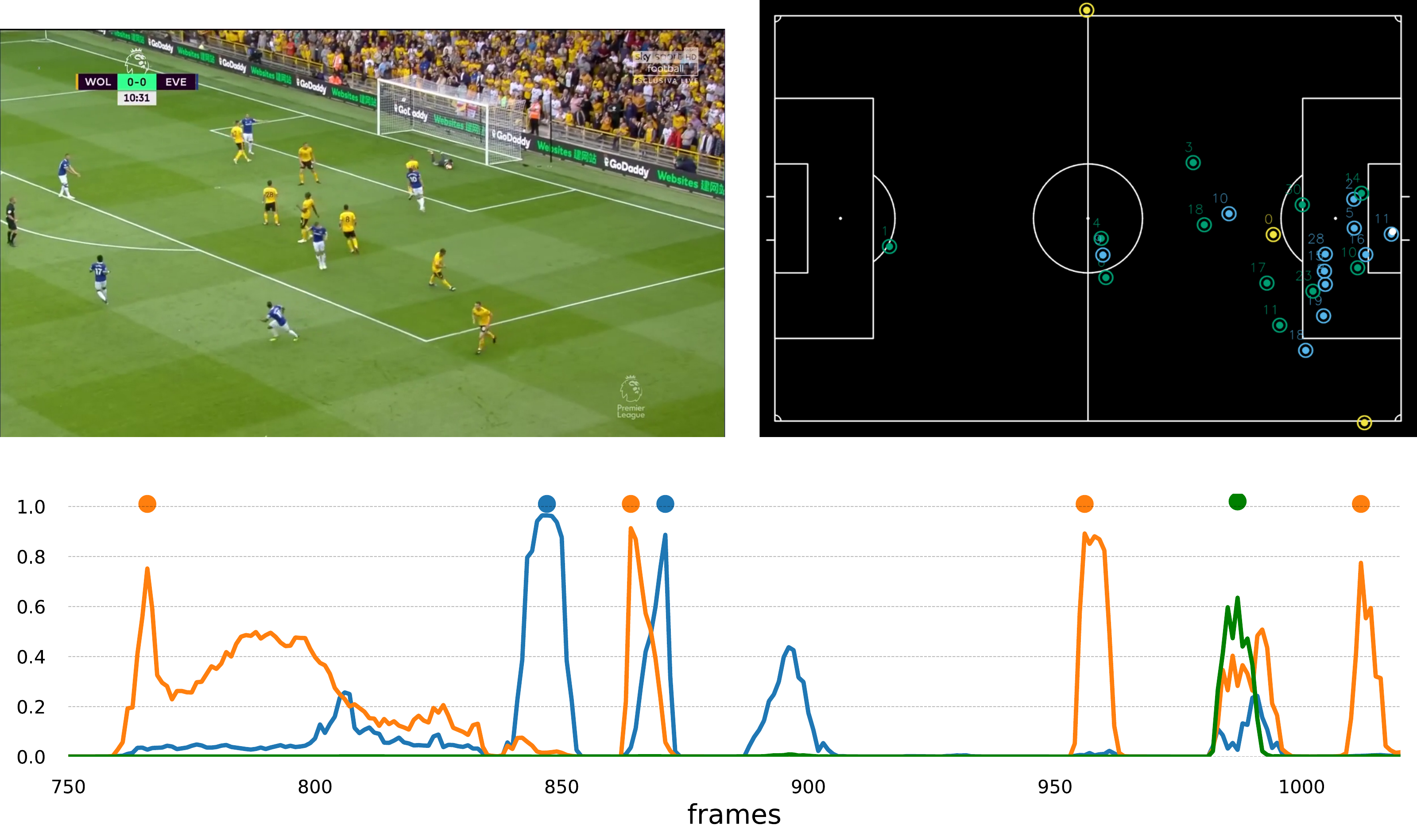}
    \caption{Pass, shot and reception detection in a soccer game: \textbf{Top-Left}: a frame from broadcast video. \textbf{Top-Right}: ball and players location rendered on a soccer pitch template. \textbf{Bottom}: model output probabilities per frame associated with passes, receptions and shots in blue, orange and green, respectively. The dots illustrate detected events after applying the event-specific threshold and non-maximum suppression process.}
    \label{fig:pass_shot_output}
\end{figure}

The majority of previous work for activity detection consider actions that have a duration \cite{chao_rethinking_2018, xu_r-c3d_2017, shou_autoloc_2018, paul_w-talc_2018,gu_ava_2018}. 
However, in the context of sport data it is difficult to obtain a large dataset that has labels for the actions and group activities along with their duration. In almost all available sport datasets, the actions and group activities are only marked at one single moment in time - when the action has happened and there is no notion of the duration of the activity. For example, a shot in soccer is identified at the frame (or timestamp) that the player kicks the ball. This type of annotation was designed to reduce ambiguity. Moreover, it is useful for sport analytics as it allows for simple metrics to be computed (e.g. counting the number of occurrences of each event), as well as more advanced metrics that encapsulate the context of play at the moment an event occurs (e.g. proximity of opposing players at the time of the shot). The lack of duration for the activities presents challenges both for training and evaluating activity detection models because losses and metrics, such as temporal intersection-over-union (temporal IOU), \cite{chao_rethinking_2018}, and insertions and deletions \cite{ward_performance_2011} cannot be used. Figure \ref{fig:pass_shot_output} illustrates the tasks we considered in this paper.

While a few studies have investigated activity recognition on sports videos \cite{roy_tora_classification_2017, fani_hockey_2017, cai2019temporal, piergiovanni2018fine}, they either considered classification of clips or detected temporal activities while localizing the start and end frames, which is not well-defined for detecting atomic events with no duration. 
Similarly, the existing body of research on utilizing trajectory data to recognize individual or group activities for team sports, such as basketball, soccer, and hockey \cite{Le2017, Lucey2015, Mehrasa2018SSAC, Wang2016SSAC}, has mainly been concerned with the player performance evaluation, not activity detection. We investigate the use of those models for group activity detection by conducting extensive experiments on a large-scale soccer dataset.

Recently, inspired by natural language processing (NLP) tasks \cite{Vaswani2017}, researchers have used transformers to implicitly model interactions between actors for action recognition \cite{girdhar_video_2019}. The transformer model relies on a self-attention mechanism to model complex interactions across different components (e.g. words in for NLP tasks or players in sports), which may be useful for predicting group activity.
This approach was extended to group activity recognition by Gavrilyuk \emph{et al.} \cite{kirill}, where the self-attention mechanism learns dynamic relationships between actors that are most important for predicting group activity, without the need to explicitly define a graph structure. This approach achieved state-of-the-art results on the volleyball \cite{ibrahim_hierarchical_2016} and collective activity dataset \cite{collective}. However, these datasets do not share the same challenges that commonly occur in soccer or hockey games, that is, the actors are relatively static in their location and there is minimal transition between events. 

In this work, we investigated the benefit of employing self-attention on the spatio-temporal embeddings extracted from ball and players trajectories as well as bounding boxes around the players to detect group activity in soccer games. To capture long-term temporal dependencies between the people for group activity detection, the trajectory-based models employ transformers on deep features learned by dilated 1D temporal convolutional network (TCN), similar to the WaveNet \cite{WaveNet} network. 
The vision-based approach utilized an inflated 3D convolution (I3D) model to capture spatio-temporal features followed by a transformer to learn interactions across players. 

More specifically, the contributions of this work are: i) conducting a comprehensive analysis on the use of visual features and trajectory data for group activity detection in sports; ii) investigation on the use of action recognition methods and extend them for group activity detection with atomic events; and iii) introducing the use of transformers for group activity detection from trajectory data. In addition, we perform extensive experiments on a large-scale soccer dataset provided by Sportlogiq from the English Premier League. The dataset contains broadcast video feeds along with group activity labels for each frame and 2D trajectories of the ball and players on real-world coordinates.

\begin{figure*}
    \centering
    \includegraphics[width=\textwidth]{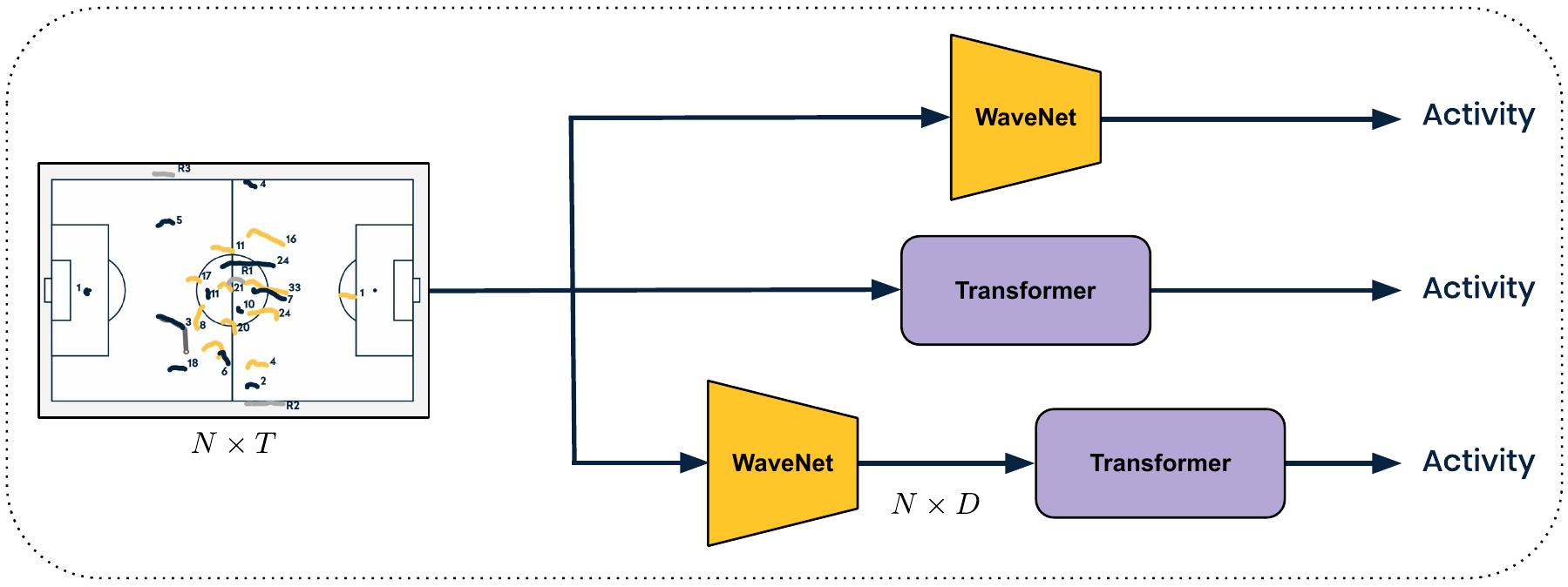}
    \caption{The trajectory-based models that are compared in this paper: the input is segments of $T$ frames including $N$ object trajectories (ball and players) for all the models. The bottom architecture extracts $D$-dimensional feature vector applied to the transformer model.}
    \label{fig:trajectory_models}
\end{figure*}

\section{Related Work}

\subsection{Trajectory-based activity recognition}
As the most widely available source of data in sports, trajectory data serves as an intermediate, sparse representation of the scene. Despite the availability of the trajectory data, there are very few research works that tried to exploit them for recognizing and detecting group activities \cite{Le2017, Lucey2015, Mehrasa2018SSAC, Wang2016SSAC}. 
In the context of sport trajectory data, the focus has been mostly to model the trajectories to either generate metrics for player and team evaluation, or to make predictions and generate trajectories for game simulations \cite{Yeh2019CVPR,Felsen2018ECCV}. In the early work, Lucey \etal used a compact representation of the trajectories by reducing the sport player trajectories into player roles to identify team formation and plays \cite{Lucey2013CVPR}. In \cite{Wang2016SSAC}, the trajectory data was converted to a binary image representing the location of the players on a template of a basketball court and then a Convolutional Neural Network (CNN) and Recurrent Neural Network (RNN) were applied to classify NBA offensive patterns. Converting the trajectories into an image representation makes the use of 2D CNN straightforward. However, the temporal aspect of the trajectory data was ignored.
In \cite{Mehrasa2018SSAC}, a two stream process was introduced to leverage both tracking and visual data. A temporal convolutional network was applied on the trajectory data to encode trajectories into an spatio-temporal embedding followed by pooling layers to aggregate the temporal dimension. Similarly, TCNs were used to learn an embedding for trajectories of the players and the ball in football to predict the outcome of a play \cite{Horton2020SSAC}. In \cite{zhong_time_2018}, hierarchical LSTMs were used for temporal encoding of the extracted features from the video frames and trajectory data followed by temporal point processes to model temporal distribution of the players in a sport game for predicting when the next activity will occur. In \cite{Felsen2018ECCV}, conditional variational autoencoders were used to generate trajectory embeddings in basketball to characterize players behaviour by conditioning on players and team identities and using heuristics to assign players to roles. In most of the previous approaches, an ordering between the players and the ball is fixed, either by enforcing relative distance to the ball or applying players roles in the game \cite{Lucey2013CVPR,Felsen2018ECCV}. Here, instead of imposing any pre-defined structure on the relationship between the players in the scene, we use transformers and leverage self-attention to fully capture the scene context.

\subsection{Vision-based activity recognition} 
Vision-based action and group activity recognition has been the subject of several studies in the literature \cite{Deng2015BMVC,wu_learning_2019,ibrahim_hierarchical_2016,QiECCV2018, AzarCVPR2019,tran_learning_2015,ji_3d_2013,simonyan_two-stream_2014}. It is commonly formulated as a classification problem over a short video segment, of one or a few seconds. Early approaches used features extracted from 2D CNNs on individual frames, followed by temporal information modeling \cite{DonahuePAMI2014, NgCVPR2015,SharmaICLR2016, donahue_long-term_2015, ibrahim_hierarchical_2016}. More recently, 3D CNNs, such as C3D and I3D models have been applied to the video segments directly to learn spatio-temporal filters \cite{ji_3d_2013, tran_learning_2015}. This approach has the advantage of extracting features from the whole scene that could be useful for predicting group activity. However, given that actions in sports videos consume a small portion of the entire frame, the extracted features may be dominated by features related to the background \cite{ibrahim_hierarchical_2016}. To overcome this limitation, recent works have begun modeling the relationship between the players in the video as this can serve as an important feature to infer group activity. Typically, this involves detecting and tracking persons in the video and pass the bounding boxes through a feature extractor. To model the relationship between the bounding boxes, many approaches have been proposed, ranging from simple heuristics, such as maximum activation over bounding boxes \cite{zhong_time_2018,roy_tora_classification_2017} and temporal information modeling with LSTMs \cite{ibrahim_hierarchical_2016}. More sophisticated approaches, such as using Graph Convolutional Neural Networks (GCN), have been introduced to model interactions between actors in a scene\cite{wu_learning_2019, wang_videos_2018}. 
One of the major difficulties in using graphs for group activities in sports is the requirement of explicitly defining a graph structure. To overcome  this limitation, Gavrilyuk \etal \cite{kirill} leveraged the self-attention mechanism in transformers to model the dynamic relationships between the actors. While this approach achieved state-of-the-art results the volleyball \cite{ibrahim_hierarchical_2016} and collective activity dataset \cite{collective}, we were interested in whether this approach would improve performance in a more dynamic setting, such as soccer.

Most methods proposed for activity detection \cite{chao_rethinking_2018, xu_r-c3d_2017, shou_autoloc_2018, paul_w-talc_2018} 
are focused on temporal localization of non-atomic actions, i.e. actions that are defined in time and have a duration such as walking and sitting \cite{gu_ava_2018}. Using temporal duration for actions makes it possible to use temporal losses such as temporal IOU \cite{chao_rethinking_2018}. However, since \emph{atomic} actions do not have a duration, these evaluation metrics cannot be used. For instance, spatio-temporal localization tasks often use 3D IOU which can only be defined for actions that span across several time steps, such as the ones in the AVA dataset \cite{gu_ava_2018}. Ward \etal \cite{ward_performance_2011} proposed to break down errors for activity recognition into more categories: insertions, deletions, fragmentation and merging. We note that these definitions are also only valid for activities that span multiple time steps. 
Another aspect that must be considered is that the number of predicted instances is important, as the end goal is to use the automatically detected actions and group activities for sport analytics purposes. For instance, predicting a pass in two consecutive frames (when a single pass is present) is an important issue for the applications of these metrics. As an example, if they are used to compute the ``number of passes under pressure for player X'', the \emph{number} of predicted instances is important.

\section{Methods}
\subsection{Trajectory-based approaches}
\label{trajectory_approach}

To investigate the effect of self-attention in group activity detection, we train three sets of models: i) a Wavenet-based \cite{WaveNet} dilated 1D TCN; ii) a transformer on the raw trajectory data; and iii) a TCN followed by a transformer, as shown in Figure \ref{fig:trajectory_models}. These are followed by a fully-connected layer to predict the activity.
For all three models, we use the ball and the players trajectories on a real-world coordinate system. Let $(x_b(t),y_b(t))$ and $\{(x_{pi}(t),y_{pi}(t))\}_{i=1}^{N-1}$ denote the location of the ball and the players at the time step $t$. Inputs are created by concatenating the location of the ball and the players within a temporal window of $T$ to form $2 \times T \times N$ feature vectors. 
During pre-processing, all the trajectories are normalized to the length and width of the corresponding stadium. We zero-pad the trajectories to deal with mis-detection and to ensure that all the inputs are of the length of $T$. 
Since the target activities are all on-the-ball actions, we ran our experiments using only the ball trajectory, and also a ball-centric representation that uses the ball and the $K$-nearest players trajectories. For the latter case, we identify the closest players using their average distance to the ball inside the temporal window.


\subsection{Vision-based approaches}

\begin{figure*}
    \centering
    \includegraphics[width=\textwidth]{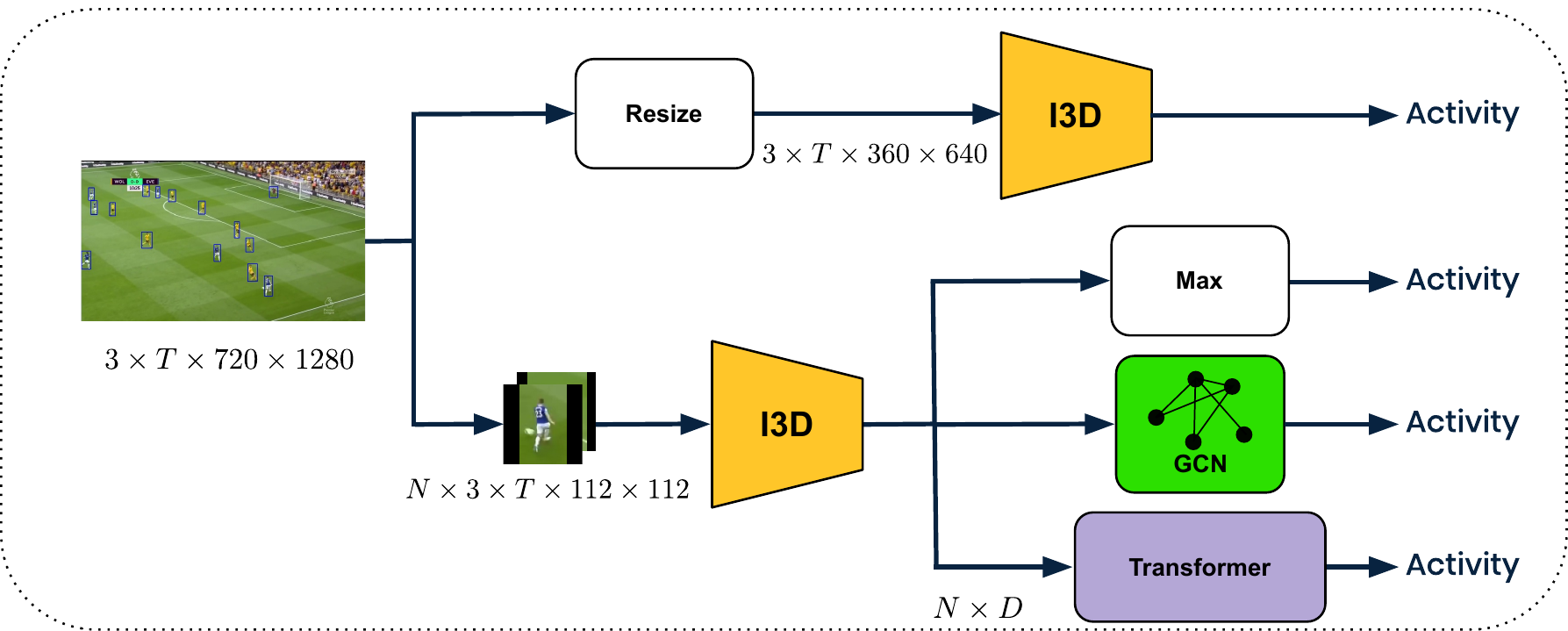}
    \caption{The vision-based methods compared on this paper. \textbf{Top}: 3D CNN applied to videos of the entire frame. \textbf{Bottom}: Tubelets are created for each of the $N$ tracked players, and a 3D CNN model is applied, resulting in a $D$-dimensional feature vector for each player. We consider a simple aggregation (max over $N$), a GCN and a transformer model.}
    \label{fig:methods}
\end{figure*}


All the vision-based models studied in this paper use an Inflated 3D CNN as the backbone \cite{carreira_quo_2017}. Inputs to the network are short clips of size $3 \times T \times H \times W$, consisting of $T$ frames of size $H \times W$, and 3 color channels. In particular, we use an inflated Resnet-18 \cite{hara_can_2017} backbone that uses 3D convolutional layers with residual connections, which outputs a 512-dimensional feature vector. Figure \ref{fig:methods} illustrates the vision-based approaches evaluated in this paper: the model on the top processes full frames, and the models on the bottom first processes tubelets from each player, and then aggregates the results.

For the I3D model used on the full frames, we consider inputs of size $3 \times T \times 360 \times 640$, and add a fully-connected layer to predict the activity.
For the models that process tubelets, we consider a collection of $N$ tracks for the players that are visible in the middle frame of the clip. We extract bounding boxes for each player in the $T$ frames of the clip, resize to a standard size of $112 \times 112$ (preserving aspect ratio), and creating one tubelet per player: a tensor of size $3 \times T \times 112 \times 112$. Tracks that are shorter than $T$ (e.g. a player entering the camera's field of view) are zero-padded. We aggregate the information from different tubeletes in three approaches: the first considers, for each class, the maximum prediction over tubelets. The second approach uses a GCN, where a graph is explicitly defined based on the distance between the players in pixel space \cite{wu_learning_2019}. The final approach uses a transformer that follows a similar formulation to \cite{kirill} by considering the set of features for all $N$ players as the sequence to be encoded. After this step, a fully-connected layer outputs a single prediction for the clip.


\section{Experiments}
\subsection{Dataset}
We conducted experiments on a dataset of 74 soccer games from the 2018-2019 English Premier League provided by Sportlogiq. The dataset includes 30FPS broadcast camera feeds at 720P resolution, the ball and all players trajectories obtained from an alternate vision-based tracking system using multiple cameras, and an event label (one of \emph{background}, \emph{pass}, \emph{reception} and \emph{shot}) at each frame.
The average number of events per game are: 912 passes, 1131 receptions and 22 shots (imbalance ratio: 41:51:1). The average number of players visible in the broadcast feed is 13 ($\pm 3.7$). A total of 64 games were used for training the models, 5 for validation and 5 for test. To have a better estimation of our model generalization capability on unseen teams and stadiums, there is no overlap between the training, validation and test teams. 

\subsection{Experimental protocol}
The vision-based models were trained on short clips of 25 frames, centered on the event frame (with $\pm 2$ frames for data augmentation). The trajectory-based models were trained on short segments of 51 frames, centered on the event frame.
For training, we included a background class, consisting of temporal windows where none of the events of interest occur, with a ratio of ~50\% of samples being background. The remaining samples are one of the events of interest. 
We trained the following models:

\begin{itemize}
    \item \textbf{Trajectory-based models}
    \begin{itemize}
        \item \textbf{TCN}: ball and players trajectories are used to train a sequence classifier consisting of a stack of Wavenet-based 1D convolutional layers. 
        \item \textbf{Transformer}: ball and players trajectories are used to train a transformer-based sequence classifier.  
        \item \textbf{Transformer on top of TCN}: ball and players trajectories are fed to  the TCN and the transformer is applied on the temporal features to classify the input sequence. 
    \end{itemize}
     \item \textbf{Vision-based models}
    \begin{itemize}
        \item \textbf{I3D trained on the whole frame}: For this model, the frames are resized to a fixed size ($640 \times 360$).
        \item \textbf{I3D trained on bounding boxes}: For the events of interest, we train the model with a spatio-temporal tubelet of the player that did the action, that is, a sequence of images from a bounding box around the player. For background events, we select a random player for training.
        \item \textbf{GCN on I3D trained on bounding boxes}: For this model, tubelets for all players in the broadcast field of view are used for training. The graph is defined based on player proximity in pixel space in the center frame.
        \item \textbf{Transformer on I3D trained on bounding boxes}: Similar to the GCN model, tubelets for all players in the broadcast field of view are used for training.
    \end{itemize}
\end{itemize}

\subsection{Evaluation metrics}
Considering the final objective of \emph{detection} performance (as opposed to \emph{classification} on clips), we evaluate the models on a sliding window of a larger segment of the game. During validation, we measured the detection performance of the models on 500-frame segments. We report the performance metrics on the test set using 15000-frame segments (8 minutes) for each game.

To evaluate the models, we first gather the predicted probabilities at each frame by centering a temporal window around it to create the input sequence. For each class $c$, a threshold  $\tau_c$ is applied on the model probabilities to identify an event. To eliminate multiple detections on a single event, a non-maximum supression (NMS) procedure is applied, with an event-specific window length $W_{\text{NMS}}$. We compute the true positives (TP), false positives (FP) and false negatives (FN) for each event as follows. A positive prediction is counted as a TP if and only if there exist a ground truth event within a temporal distance of $W_{\text{eval}}$ frames; otherwise it will be a FP detection. If there are multiple ground truth events exist within the temporal window, the positive prediction is assigned to the nearest ground truth event. A ground truth event is counted as a FN if there are no positive predictions assigned to it. During our experiments, we set the evaluation window length to 51 frames, which is less than one second around an actual event.

We optimized the hyperparameters $\tau_c$ and $W_{\text{NMS}}$ to achieve the maximum F-score on validation segments that were randomly chosen from 5 games in our validation dataset. We performed a grid search with values of $\tau_c$ ranging from $0.3$ to $0.98$ with the step of $0.02$, and  $W_{\text{NMS}}$ ranging from $3$ to $59$ frames with the step of $2$ frames. 

We evaluated the models on the test set in terms of precision, recall and F-score computed at $\tau_c^*$ and $W_{\text{NMS}}^*$. We also measured the temporal distance (TD) error,  as the time difference (in seconds) between the ground truth and the corresponding predicted frame, for all the true positive predictions of each event. Their values at 0.5 and 0.95 percentiles were used as the evaluation metrics. This evaluation procedure is designed to cope with the fact that the events of interest are \emph{atomic} (they are labelled for a single frame), but there is an uncertainty in their labelling (annotation may be incorrect by a few frames). It also does not penalize the model if it predicts the event within a small window of the actual frame. The precision of the prediction in time is captured by the TD metric.

\section{Results and Discussion}
All vision and trajectory-based models were trained and tested for the same games. The models were initially trained for action recognition and were evaluated to detect the same events in time.

\subsection{Trajectory-based models}
Using the ball and the players trajectories, we trained the three models in Section \ref{trajectory_approach} on the training set. 
Since the target activities are all on-the-ball actions, we ran our experiments using the trajectories of the $K$-nearest players to the ball. This helps all of the models   attend to the region around the ball, which is where the action is happening.
Table \ref{tbl:trajectory_5player_result} shows the detection precision, recall, F-score, $0.5$ and $0.95$ percentiles of temporal distance error. 
According to the results, the combination of TCN and transformer improves the activity detection performance for all of the desired events. Moreover, the events are detected more accurately in time (lower $0.5$ and $0.95$ percentiles of TD errors), suggesting the usefulness of employing self-attention mechanism on top of feature extractors. 

We evaluated the best-performing model (TCN+Transfomer) in a series of scenarios, that are summarized in Table \ref{tbl:trajectory_ablation}.
First, we report the performance of the model using the ball and all players trajectories. The result suggests that the model does not benefit from accessing the trajectories of all players and limiting the input data to the region around the ball helps the detection performance.
Second, to further assess the impact of the players trajectories, we re-trained the TCN+Transformer model using only the ball trajectory. The model's performance suggests that the ball trajectory plays an important role in detecting on-the-ball group activities. 
This also highlights a shortcoming of the trajectory-based models, which is detecting passes and shots that are blocked shortly after the attempt has been made. In these situation, the model cannot perceive any ball movements and is unable to detect and identify the correct action.

\begin{figure}
    \centering
    \includegraphics[width=0.8\columnwidth]{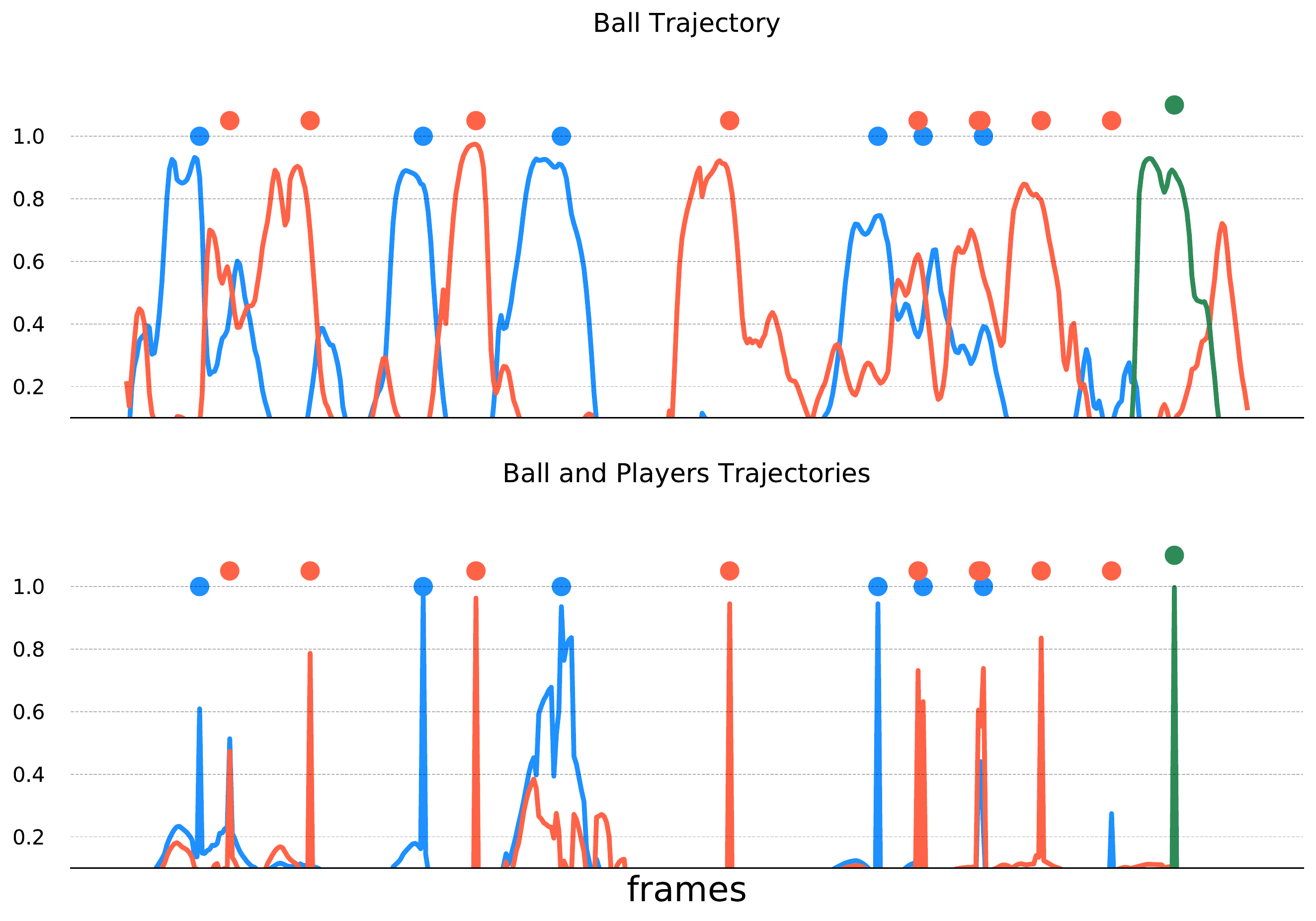}
    \caption{How TCN with transformer behaves in detecting our desired events using \textbf{Top}: only ball trajectory data, \textbf{Bottom}: ball and 5-closest players data. Model output probabilities per frame associated with pass, reception and shot in blue, orange and green, respectively. The dots indicate ground truth events.}
    \label{fig:model_behavior}
\end{figure}

Figure \ref{fig:model_behavior} illustrates how the model behaves differently with and without having access to the players trajectories. The figure shows the prediction probabilities for the desired events, versus the ground-truth during a 15-second segment.
While both models were able to detect the majority of the events, using the players trajectories helps increasing the model's temporal resolution in detecting consecutive actions. It is worth mentioning that without using the players trajectories, the optimised NMS window lengths for passes, receptions and shots are 3.7, 1.2 and 1.9 times higher.

Finally, we fine-tuned the pre-trained TCN+Transformer model using partially observed tracking data obtained from the broadcast feed. While having shorter tracks, the partial broadcast trajectories are mostly available for the ball and the area where the activities are occurring. The results suggest that it is possible to achieve comparable detection performance using trajectories obtained by tracking players in broadcast videos.

\begin{table}
\begin{center}
\begin{adjustbox}{width=\columnwidth}
\begin{tabular}{l|l|ccccc}
 \textbf{Model} & \textbf{Event} & \textbf{Precision} & \textbf{Recall} & \textbf{F-Score} & \textbf{TD} @ $0.5$ & \textbf{TD} @ $0.95$\\ \midrule
  \multirow{3}{*}{TCN} & Pass & $0.60$ & $0.50$ & $0.55$ & $0.28$ & $0.86$ \\ 
 & Reception & $0.51$ & $0.62$ & $0.56$ & $0.38$ & $0.87$\\ 
 & Shot  & $0.51$ & $0.62$ & $0.56$ & $0.38$ & $0.87$ \\ \midrule
  \multirow{3}{*}{Transformer} & Pass & $0.82$ & $0.89$ & $0.86$ & $0.20$ & $0.50$ \\ 
 & Reception & $0.74$ & $0.88$ & $0.80$ & $0.18$ & $0.52$\\ 
 & Shot  & $0.43$ & $0.88$ & $0.58$ & $0.10$ & $0.12$ \\ \midrule
  \multirow{3}{*}{\specialcell{TCN + \\Transformer}} & Pass & $0.87$ & $0.86$ & $0.87$ & $0.20$ & $0.48$\\
 & Reception & $0.80$ & $0.86$ & $0.83$ & $0.18$ & $0.50$\\ 
 & Shot  & $0.60$ & $0.88$ & $0.71$ & $0.19$ & $0.19$ \\ 
\end{tabular}
\end{adjustbox}
\end{center}
\caption{Performance metrics computed for trajectory-based models applied on the test dataset using 2D positions of the ball and 5-closest players to the ball.}
\label{tbl:trajectory_5player_result}
\end{table}

\begin{table}
\begin{center}
\begin{adjustbox}{width=\columnwidth}
\begin{tabular}{l|l|ccccc}
 \textbf{Model} & \textbf{Event} & \textbf{Precision} & \textbf{Recall} & \textbf{F-Score} & \textbf{TD} @ $0.5$ & \textbf{TD} @ $0.95$\\ \midrule
  \multirow{3}{*}{\specialcell{Complete Trajectories \\Ball and All Players}} & Pass & $0.71$ & $0.92$ & $0.8$ & $0.28$ & $0.66$\\
 & Reception & $0.82$ & $0.79$ & $0.8$ & $0.24$ & $0.56$\\ 
 & Shot  & $0.71$ & $0.63$ & $0.67$ & $0.18$ & $0.18$ \\ \midrule
 \multirow{3}{*}{\specialcell{Complete Trajectories \\Ball only}} & Pass & $0.81$ & $0.77$ & $0.78$ & $0.28$ & $0.5$\\
 & Reception & $0.79$ & $0.79$ & $0.79$ & $0.22$ & $0.52$\\ 
 & Shot  & $0.54$ & $0.88$ & $0.67$ & $0.18$ & $0.21$ \\ \midrule
  \multirow{3}{*}{\specialcell{Partial Broadcast Trajectories \\Ball and 5-closest Players}} & Pass & $0.80$ & $0.88$ & $0.84$ & $0.27$ & $0.62$\\
 & Reception & $0.86$ & $0.73$ & $0.79$ & $0.24$ & $0.56$\\ 
 & Shot  & $0.83$ & $0.63$ & $0.71$ & $0.08$ & $0.08$ \\ 
\end{tabular}
\end{adjustbox}
\end{center}
\caption{Ablation analysis of the trajectory-based models. We used the TCN+Transformer model for all the experiments.}
\label{tbl:trajectory_ablation}
\end{table}

\begin{figure}
    \centering
    \includegraphics[width=0.9\columnwidth]{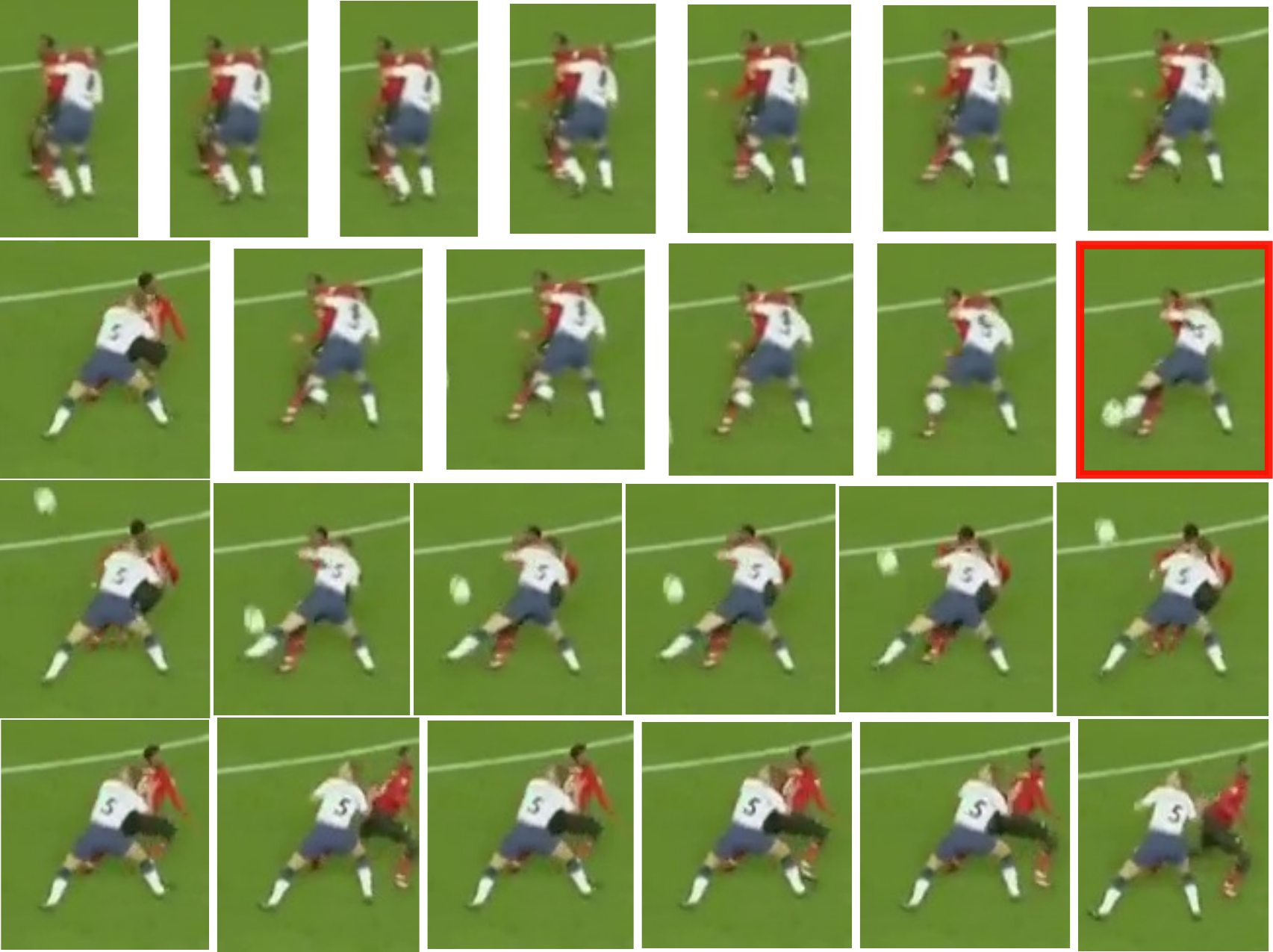}
    \caption{Example of a common error scenario for all models investigated in this work. We show 12 frames before and after a predicted event (indicated with a red border). Without any further context, it appears that player with jersey number 5 is making a pass, while he is actually taking possession of the ball.}
    \label{fig:pass_hard}
\end{figure}

\subsection{Vision-based models}

Table \ref{tbl:vision_result} reports the results for the vision-based models. The I3D model trained on full-frame information performed best for all three actions evaluated in this paper. We also observed that the Transformer model improved results over I3D trained on bounding boxes for passes and receptions, but obtained worse results for shots. Surprisingly, the model using GCN to capture the interactions between players performed worse. One of the reason can be the way the graph is defined which is based on players' location proximity to the center of the frame.   

Overall, we noticed that all models had errors in situations where a player receives the ball and takes a long period of time to control it. This is illustrated in Figure \ref{fig:pass_hard}: over this segment (12 future frames), it appears that the player is making a pass, while by observing 2 more seconds in future we observe that the same player took possession of the ball. This is a limitation given by the short length of the segments used for training/testing. 

Comparing the I3D frame and I3D BBoxes + Transformer models more closely, we see that their detection performance for pass and reception was reasonably close, while for shots, the model trained on bounding boxes performed much worse due to its very low precision. A visual analysis of the errors showed that this model struggles to distinguish between passes and shots made close to the goal area - since only visual information around players is used, the model has no access to the trajectory of the ball.

\begin{table}
\begin{center}
\begin{adjustbox}{width=\columnwidth}
\begin{tabular}{l|l|ccccc}

 \textbf{Model} & \textbf{Event} & \textbf{Precision} & \textbf{Recall} & \textbf{F-Score} & \textbf{TD} @ $0.5$ & \textbf{TD} @ $0.95$\\ \midrule

\multirow{3}{*}{I3D Frame}  & Pass      & 0.78 & 0.91 & 0.84 & 0.03 & 0.18 \\
                             & Reception & 0.74 & 0.81 & 0.77 & 0.04 & 0.41 \\
                            & Shot      & 0.83 & 0.63 & 0.71 & 0.13 & 0.13 \\ \midrule
\multirow{3}{*}{I3D BBoxes} & Pass      & 0.79 & 0.82 & 0.81 & 0.15 & 0.38 \\
                            & Reception & 0.69 & 0.50  & 0.73 & 0.02 & 0.02 \\
                            & Shot      & 0.40  & 0.50  & 0.44 & 0.09 & 0.45 \\ \midrule
\multirow{3}{*}{\specialcell{I3D BBoxes +\\ GCN}} & Pass      & 0.80  & 0.62 & 0.70  & 0.09 & 0.43   \\
&Reception & 0.64 & 0.47 & 0.54 & 0.08 & 0.47   \\
&Shot      & 0.01    & 0.38 & 0.01 & 0.36 & 0.36   \\ \midrule

\multirow{3}{*}{\specialcell{I3D BBoxes +\\ Transformer}} & Pass      & 0.79 & 0.89 & 0.83 & 0.09 & 0.29 \\
        & Reception & 0.70  & 0.78 & 0.74 & 0.07 & 0.44 \\
        & Shot      & 0.29 & 0.63 & 0.40  & 0.01 & 0.01
\end{tabular}
\end{adjustbox}
\end{center}
\caption{Performance metrics computed for vision-based models applied on the test dataset.}
\label{tbl:vision_result}
\end{table}

Given results obtained from both the trajectory and vision based models, we observe that most events can be detected using either vision or trajectory-based  approaches with a temporal resolution of less than $0.5$ seconds. We note that the trajectory-based model performs better in detecting certain events that depend on the physical location in the pitch (e.g. for shots the model need to infer that the ball is moving toward the goal). We also note that fusion between the trajectory and visual streams can be considered as one of the potential extension of the current work. 

\section{Conclusion}

In this paper, we conducted a comprehensive analysis on the use of vision- and trajectory-based methods for group activity detection on a large-scale soccer dataset, and introduced the use of transformers applied on temporal feature representations for group activity detection. We focused on detecting \emph{pass}, \emph{shot} and \emph{reception} in soccer games, and showed that models trained for activity recognition on clips can perform well for the \emph{detection} task, although with some limitations, such as not recognizing sequences that require a longer temporal context (e.g. as shown in Figure \ref{fig:pass_hard}).
Training models for activity detection is a promising way to address this issue, but requires new models that are well defined for actions without a duration.

We also observed that the I3D models trained on the whole frame captured by the broadcast camera performed better than directly modeling player interactions using transformers or GCNs. However, we hypothesize that this needs deeper investigation for static camera setups, since a large fraction of a frame is background for which an attention mechanism might be required to mitigate this issue. 


{\small
\bibliographystyle{ieee_fullname}
\bibliography{biblio}

\begin{thebibliography}{10}\itemsep=-1pt

\bibitem{AzarCVPR2019}
Sina~Mokhtarzadeh Azar, Mina~Ghadimi Atigh, Ahmad Nickabadi, and Alexandre
  Alahi.
\newblock Convolutional relational machine for group activity recognition.
\newblock In {\em IEEE Conference on Computer Vision and Pattern Recognition
  (CVPR)}, pages 7892--7901, 2019.

\bibitem{cai2019temporal}
Zixi Cai, Helmut Neher, Kanav Vats, David~A Clausi, and John Zelek.
\newblock Temporal hockey action recognition via pose and optical flows.
\newblock In {\em IEEE Conference on Computer Vision and Pattern Recognition
  Workshops (CVPRW)}, pages 0--0, 2019.

\bibitem{carreira_quo_2017}
Joao Carreira and Andrew Zisserman.
\newblock Quo vadis, action recognition? a new model and the kinetics dataset.
\newblock In {\em IEEE Conference on Computer Vision and Pattern Recognition
  (CVPR)}, pages 6299--6308, 2017.

\bibitem{chao_rethinking_2018}
Yu-Wei Chao, Sudheendra Vijayanarasimhan, Bryan Seybold, David~A Ross, Jia
  Deng, and Rahul Sukthankar.
\newblock Rethinking the faster {R-CNN} architecture for temporal action
  localization.
\newblock In {\em IEEE Conference on Computer Vision and Pattern Recognition
  (CVPR)}, pages 1130--1139, 2018.

\bibitem{Deng2015BMVC}
Zhiwei Deng, Mengyao Zhai, Lei Chen, Yuhao Liu, Srikanth Muralidharan,
  Mehrsan~Javan Roshtkhari, and Greg Mori.
\newblock Deep structured models for group activity recognition.
\newblock In {\em British Machine Vision Conference (BMVC)}, 2016.

\bibitem{donahue_long-term_2015}
Jeffrey Donahue, Lisa Anne~Hendricks, Sergio Guadarrama, Marcus Rohrbach,
  Subhashini Venugopalan, Kate Saenko, and Trevor Darrell.
\newblock Long-term recurrent convolutional networks for visual recognition and
  description.
\newblock In {\em IEEE Conference on Computer Vision and Pattern Recognition
  (CVPR)}, pages 2625--2634, 2015.

\bibitem{DonahuePAMI2014}
Jeff Donahue, Lisa~Anne Hendricks, Marcus Rohrbach, Subhashini Venugopalan,
  Sergio Guadarrama, Kate Saenko, and Trevor Darrell.
\newblock Long-term recurrent convolutional networks for visual recognition and
  description.
\newblock {\em IEEE Transactions on Pattern Analysis and Machine Intelligence},
  39:677--691, 2014.

\bibitem{fani_hockey_2017}
Mehrnaz Fani, Helmut Neher, David~A Clausi, Alexander Wong, and John Zelek.
\newblock Hockey action recognition via integrated stacked hourglass network.
\newblock In {\em IEEE Conference on Computer Vision and Pattern Recognition
  Workshops (CVPRW)}, pages 29--37, 2017.

\bibitem{Felsen2018ECCV}
Panna Felsen, Patrick Lucey, and Sujoy Ganguly.
\newblock Where will they go? predicting fine-grained adversarial multi-agent
  motion using conditional variational autoencoders.
\newblock In {\em European Conference on Computer Vision (ECCV)}, pages
  732--747, 2018.

\bibitem{kirill}
Kirill Gavrilyuk, Ryan Sanford, Mehrsan Javan, and Cees Snoek.
\newblock Actor-transformers for group activity recognition.
\newblock In {\em IEEE Conference on Computer Vision and Pattern Recognition
  (CVPR)}, 2020.

\bibitem{girdhar_video_2019}
Rohit Girdhar, Joao Carreira, Carl Doersch, and Andrew Zisserman.
\newblock Video action transformer network.
\newblock In {\em IEEE Conference on Computer Vision and Pattern Recognition
  (CVPR)}, pages 244--253, 2019.

\bibitem{gu_ava_2018}
Chunhui Gu, Chen Sun, David~A Ross, Carl Vondrick, Caroline Pantofaru, Yeqing
  Li, Sudheendra Vijayanarasimhan, George Toderici, Susanna Ricco, Rahul
  Sukthankar, et~al.
\newblock {AVA}: A video dataset of spatio-temporally localized atomic visual
  actions.
\newblock In {\em IEEE Conference on Computer Vision and Pattern Recognition
  (CVPR)}, pages 6047--6056, 2018.

\bibitem{hara_can_2017}
Kensho Hara, Hirokatsu Kataoka, and Yutaka Satoh.
\newblock Can spatiotemporal {3D} {CNNs} retrace the history of {2D} {CNNs} and
  {ImageNet}?
\newblock In {\em IEEE Conference on Computer Vision and Pattern Recognition
  (CVPR)}, pages 6546--6555, 2018.

\bibitem{Horton2020SSAC}
Michael Horton.
\newblock Learning feature representations from football tracking.
\newblock In {\em MIT Sloan Sports Analytics Conference}, 2020.

\bibitem{ibrahim_hierarchical_2016}
Mostafa~S Ibrahim, Srikanth Muralidharan, Zhiwei Deng, Arash Vahdat, and Greg
  Mori.
\newblock A hierarchical deep temporal model for group activity recognition.
\newblock In {\em IEEE Conference on Computer Vision and Pattern Recognition
  (CVPR)}, pages 1971--1980, 2016.

\bibitem{ji_3d_2013}
Shuiwang Ji, Wei Xu, Ming Yang, and Kai Yu.
\newblock {3D} convolutional neural networks for human action recognition.
\newblock {\em IEEE Transactions on Pattern Analysis and Machine Intelligence},
  35(1):221--231, 2012.

\bibitem{Keane2019SSAC}
Evin Keane, Philippe Desaulniers, Luke Bornn, and Mehrsan Javan.
\newblock Data-driven lowlight and highlight reel creation based on explainable
  temporal game models.
\newblock In {\em MIT Sloan Sports Analytics Conference}, 2019.

\bibitem{Le2017}
Hoang~M Le, Peter Carr, Yisong Yue, and Patrick Lucey.
\newblock Data-driven ghosting using deep imitation learning.
\newblock In {\em MIT Sloan Sports Analytics Conference}, 2017.

\bibitem{Lucey2013CVPR}
Patrick Lucey, Alina Bialkowski, Peter Carr, Stuart Morgan, Iain Matthews, and
  Yaser Sheikh.
\newblock Representing and discovering adversarial team behaviors using player
  roles.
\newblock In {\em IEEE Conference on Computer Vision and Pattern Recognition
  (CVPR)}, pages 2706--2713, 2013.

\bibitem{Lucey2015}
P. Lucey, A. Bialkowski, M. Monfort, P. Carr, and I. Matthews.
\newblock Quality vs quantity: Improved shot prediction in soccer using
  strategic features from spatiotemporal data.
\newblock In {\em MIT Sloan Sports Analytics Conference}, 2015.

\bibitem{Mehrasa2018SSAC}
Nazanin Mehrasa, Yatao Zhong, Frederick Tung, Luke Bornn, and Greg Mori.
\newblock Deep learning of player trajectory representations for team activity
  analysis.
\newblock In {\em MIT Sloan Sports Analytics Conference}, 2018.

\bibitem{NgCVPR2015}
Joe Yue-Hei Ng, Matthew~J. Hausknecht, Sudheendra Vijayanarasimhan, Oriol
  Vinyals, Rajat Monga, and George Toderici.
\newblock Beyond short snippets: Deep networks for video classification.
\newblock In {\em IEEE Conference on Computer Vision and Pattern Recognition
  (CVPR)}, 2015.

\bibitem{WaveNet}
Aaron van~den Oord, Sander Dieleman, Heiga Zen, Karen Simonyan, Oriol Vinyals,
  Alex Graves, Nal Kalchbrenner, Andrew Senior, and Koray Kavukcuoglu.
\newblock Wavenet: A generative model for raw audio.
\newblock {\em arXiv preprint arXiv:1609.03499}, 2016.

\bibitem{paul_w-talc_2018}
Sujoy Paul, Sourya Roy, and Amit~K Roy-Chowdhury.
\newblock W-{TALC}: {Weakly}-supervised temporal activity localization and
  classification.
\newblock In {\em European Conference on Computer Vision (ECCV)}, pages
  563--579, 2018.

\bibitem{piergiovanni2018fine}
AJ Piergiovanni and Michael~S Ryoo.
\newblock Fine-grained activity recognition in baseball videos.
\newblock In {\em IEEE Conference on Computer Vision and Pattern Recognition
  Workshops (CVPRW)}, pages 1740--1748, 2018.

\bibitem{QiECCV2018}
Mengshi Qi, Jie Qin, Annan Li, Yunhong Wang, Jiebo Luo, and Luc Van~Gool.
\newblock stagnet: An attentive semantic {RNN} for group activity recognition.
\newblock In {\em European Conference on Computer Vision (ECCV)}, pages
  101--117, 2018.

\bibitem{Schulte2017DMKD}
Oliver Schulte, Mahmoud Khademi, Sajjad Gholami, Zeyu Zhao, Mehrsan Javan, and
  Philippe Desaulniers.
\newblock A markov game model for valuing actions, locations, and team
  performance in ice hockey.
\newblock {\em Data Mining and Knowledge Discovery}, 31(6):1735--1757, 2017.

\bibitem{Schulte2017SSAC}
Oliver Schulte, Zeyu Zhao, Mehrsan Javan, and Philippe Desaulniers.
\newblock Apples-to-apples: Clustering and ranking nhl players using location
  information and scoring impact.
\newblock In {\em MIT Sloan Sports Analytics Conference}, 2017.

\bibitem{SharmaICLR2016}
Shikhar Sharma, Ryan Kiros, and Ruslan Salakhutdinov.
\newblock Action recognition using visual attention.
\newblock In {\em International Conference on Learning Representations (ICLR)
  Workshops}, 2016.

\bibitem{shou_autoloc_2018}
Zheng Shou, Hang Gao, Lei Zhang, Kazuyuki Miyazawa, and Shih-Fu Chang.
\newblock {AutoLoc}: {Weakly}-supervised temporal action localization in
  untrimmed videos.
\newblock In {\em European Conference on Computer Vision (ECCV)}, pages
  154--171, 2018.

\bibitem{simonyan_two-stream_2014}
Karen Simonyan and Andrew Zisserman.
\newblock Two-stream convolutional networks for action recognition in videos.
\newblock In {\em Advances in Neural Information Processing Systems (NIPS)},
  pages 568--576. 2014.

\bibitem{roy_tora_classification_2017}
Moumita~Roy Tora, Jianhui Chen, and James~J Little.
\newblock Classification of puck possession events in ice hockey.
\newblock In {\em IEEE conference on Computer Vision and Pattern Recognition
  Workshops (CVPRW)}, pages 147--154, 2017.

\bibitem{tran_learning_2015}
Du Tran, Lubomir Bourdev, Rob Fergus, Lorenzo Torresani, and Manohar Paluri.
\newblock Learning spatiotemporal features with 3d convolutional networks.
\newblock In {\em IEEE international Conference on Computer Vision (ICCV)},
  pages 4489--4497, 2015.

\bibitem{Vaswani2017}
Ashish Vaswani, Noam Shazeer, Niki Parmar, Jakob Uszkoreit, Llion Jones,
  Aidan~N Gomez, {\L}ukasz Kaiser, and Illia Polosukhin.
\newblock Attention is all you need.
\newblock In {\em Advances in Neural Information Processing Systems (NIPS)},
  pages 5998--6008, 2017.

\bibitem{Wang2016SSAC}
Kuan-Chieh Wang and Richard Zemel.
\newblock Classifying {NBA} offensive plays using neural networks.
\newblock In {\em MIT Sloan Sports Analytics Conference}, volume~4, 2016.

\bibitem{wang_videos_2018}
Xiaolong Wang and Abhinav Gupta.
\newblock Videos as space-time region graphs.
\newblock In {\em European conference on computer vision (ECCV)}, pages
  399--417, 2018.

\bibitem{ward_performance_2011}
Jamie~A Ward, Paul Lukowicz, and Hans~W Gellersen.
\newblock Performance metrics for activity recognition.
\newblock {\em ACM Transactions on Intelligent Systems and Technology (TIST)},
  2(1):1--23, 2011.

\bibitem{collective}
{Wongun Choi}, K. {Shahid}, and S. {Savarese}.
\newblock What are they doing? : Collective activity classification using
  spatio-temporal relationship among people.
\newblock In {\em 2009 IEEE 12th International Conference on Computer Vision
  Workshops, ICCV Workshops}, pages 1282--1289, 2009.

\bibitem{wu_learning_2019}
Jianchao Wu, Limin Wang, Li Wang, Jie Guo, and Gangshan Wu.
\newblock Learning actor relation graphs for group activity recognition.
\newblock In {\em IEEE Conference on Computer Vision and Pattern Recognition
  (CVPR)}, pages 9964--9974, 2019.

\bibitem{xu_r-c3d_2017}
Huijuan Xu, Abir Das, and Kate Saenko.
\newblock R-{C3D}: {Region} convolutional {3D} network for temporal activity
  detection.
\newblock In {\em IEEE International Conference on Computer Vision (ICCV)},
  pages 5783--5792, 2017.

\bibitem{Yeh2019CVPR}
Raymond~A Yeh, Alexander~G Schwing, Jonathan Huang, and Kevin Murphy.
\newblock Diverse generation for multi-agent sports games.
\newblock In {\em IEEE Conference on Computer Vision and Pattern Recognition
  (CVPR)}, pages 4610--4619, 2019.

\bibitem{zhong_time_2018}
Yatao Zhong, Bicheng Xu, Guang-Tong Zhou, Luke Bornn, and Greg Mori.
\newblock Time perception machine: Temporal point processes for the when, where
  and what of activity prediction.
\newblock {\em arXiv preprint arXiv:1808.04063}, 2018.

\end{thebibliography}
}

\end{document}